# Resetting a fixed broken ELBO

R. I. Cukier

*Department of Chemistry, Michigan State University, East Lansing, Michigan 48824-1322, USA*

cukier@chemistry.msu.edu

Abstract

Variational autoencoders (VAEs) are one class of generative probabilistic latent-variable models designed for inference based on known data. They balance reconstruction and regularizer terms. A variational approximation produces an evidence lower bound (ELBO). Multiplying the regularizer term by $\beta$ provides a $\beta$-VAE/ELBO, improving disentanglement of the latent space. However, any $\beta \neq 1$ value violates the laws of conditional probability. To provide a similarly-parameterized VAE, we develop a Rényi (versus Shannon) entropy VAE, and a variational approximation RELBO that introduces a similar parameter. The Rényi VAE has an additional Rényi regularizer-like term with a conditional distribution that is not learned. The term is evaluated essentially analytically using a Singular Value Decomposition method.

## 1. Introduction

Generative Adversarial Networks (GANs) [1, 2] and Variational Autoencoders (VAEs) [3,4] are two generative probabilistic models introduced with the advent of deep neural networks (DNNs). They learn the probability density function (pdf) $p(x)$ of a distribution $P_X$ ($x \in \mathcal{X}_X$) from identically, independently distributed (i.i.d.) data $x_i$ ($i=1,2,...,N$) of $N_x$-dimensional data. Once $p(x)$ is learned, new instances can be readily generated. An often-used strategy introduces a latent space probability distribution $p(y)$ ($y \in \mathcal{X}_Y$), of dimension $N_y$ with $N_y \ll N_x$, expressing the notion of much redundant information in the high-dimensional $X$ space.

VAEs rely on an identity based on the conditionals $p(x|y)$ and $p(y|x)$ of a joint pdf $p(x,y)$. The approach leads to an evidence lower bound (ELBO) on the data pdf $p(x)$ that balances reconstruction (autoencoder) and regularizer terms. ELBO-based ML approaches have been quite successful in generative mode.[3,4,5] However, the VAE has difficulty with disentangling the latent factors in unsupervised data, limiting its use. Disentangling [6,7] refers to improving the ability of the latent space to focus on particular aspects of the data. To improve the original VAE, a $\beta$-VAE was introduced [6,7] that simply multiplies the regularizer by a numerical factor, $\beta$>1, improving the efficacy of the latent space. In a related vein, the information bottleneck approach of Tishby [8] can be formulated as a VAE where terms are balanced with a $\beta$ factor [9]. $\beta$-VAE is also related [10, 7] to rate distortion theory. [11] This view emphasizes the interpolation between $\beta \ll 1$ corresponding to high rate/low distortion, essentially autoencoding, and $\beta \gg 1$ with high distortion/low rate. Similar in spirit is infoVAE [12] that, in addition to a $\beta$ balance term, adds a parameterized mutual information between data and latent spaces.

These approaches do improve the disentanglement of the latent space. However, as is evident from the form of the $\beta$-VAE expression, if one asserts that the correct joint distribution has been variationally obtained it leads to the identity

$$H(P_{X|Y}) + \beta I(P_X; P_Y) = H(P_X) + (\beta - 1) I(P_X; P_Y) \tag{1.1}$$

in terms of entropy $H(P_X)$, conditional entropy $H(P_{X|Y})$ and mutual information $I(P_X, P_Y)$.[11] In Eq. (1.1) $H(P_{X|Y})$ [$\beta I(P_X; P_Y)$] corresponds to the reconstruction [regularizer] term. Unless $\beta = 1$, the result depends on the latent-space distribution $P_Y$ and thus violates the laws of conditional probability where, by definition, $I(P_X; P_Y) \doteq H(P_X) - H(P_{X|Y})$. Appendix A elucidates this problem from the $\beta$-VAE perspective. Thus, the use of (Gibbs)-Shannon entropy

$$H_1(P_X) \doteq \sum_{x \in \mathcal{X}_x} p(x)(-\log p(x)) \tag{1.2}$$

to construct a VAE of this variety is not consistent and, if incorporating an additional parameter to balance reconstruction and regularizer terms is desired, another version of entropy is required.

Kolmogorov [13] axiomatized probability theory and suggested that

$$f^{-1}\left[\sum_{x \in \mathcal{X}_x} p(x) f(-\log p(x))\right] \tag{1.3}$$

also serves as an entropy. Rényi [14] showed that the linear or exponential choices, $f(x) = x$ or $f(x) = 2^{(1-\alpha)x}$, lead to additive entropies. The former rule produces Shannon's entropy $H_1(P_X)$ in Eq. (1.2) and the latter Rényi's $H_\alpha(P_X)$ entropy, with explicit form

$$H_\alpha(P_X) \doteq \frac{1}{1-\alpha} \log \sum_{x \in \mathcal{X}_x} p(x)^\alpha; \quad (\alpha > 0, \ \alpha \neq 1). \tag{1.4}$$

Rényi also introduced a divergence

$$D_\alpha[P_X \| Q_X] \doteq \frac{1}{\alpha - 1} \log \sum_{x \in \mathcal{X}_x} p^\alpha(x) q^{1-\alpha}(x); \quad (\alpha > 0, \ \alpha \neq 1). \tag{1.5}$$

Properties of the Rényi divergence have been summarized. [15] The $\alpha \to 1$ limit of Eq. (1.5) is the Kullback-Liebler divergence

$$D[P_X \| Q_X] \doteq \sum_{x \in \mathcal{X}_X} p(x) \log(p(x)/q(x)). \tag{1.6}$$

In this work, we use Rényi's entropy and divergence, generalized to joint/conditional pdfs, to obtain a Rényi ELBO (RELBO) version of a VAE:

$$\log p(x) \geq \underbrace{\mathbb{E}_{y \sim V(y|x)} \log \tilde{V}(x|y)}_{\text{reconstruction}} - \underbrace{\frac{1}{\alpha} D[V(y|x) \| q(y)]}_{\text{regularizer}} + \underbrace{\frac{1-\alpha}{\alpha} D_\alpha [W(y|x) \| q(y)]}_{\text{Renýi regularizer}}. \tag{1.7}$$

In Eq. (1.7), the true joint data $X$ - latent $Y$ distribution is denoted as $P_{XY} \doteq W_{Y|X} P_X$ with density $p(x,y) \doteq W(y|x) p(x)$. $V(y|x)$ [$\tilde{V}(x|y)$] is a parameterized encoder [decoder] and $q(y)$ a prior. The divergences are with respect to latent variable, $Y$. The RELBO has reconstruction and regularizer terms supplemented by a "Rényi regularizer" term that involves the true conditional distribution $W(y|x)$ to be modeled (versus learned). For $\alpha \to 1$ the RELBO reduces to the "SELBO" (Shannon ELBO) VAE:

$$\log p(x) \geq \mathbb{E}_{z \sim V(y|x)} \log \tilde{V}(x|y) - D[V(y|x) \| q(y)] \tag{1.8}$$

The additional term in RELBO vs. SELBO has a different character as it incorporates the true conditional $W(y|x)$.

To obtain a RELBO expression several issues must be addressed. First, in Section 2, we provide a definition of joint Rényi entropy and a consistent generalization of the Shannon joint Kullback-Leiber divergence to a Rényi joint divergence and mutual information, as summarized in Appendix B. This expression is used in Section 3 to obtain, via mutual information expressions, the Rényi VAE expression in Eq. (1.7). The true conditional $W(y|x)$ is modeled with probabilistic PCA (P-PCA) introduced by Roweis.[16,17] With this essentially Gaussian model in hand, Section 4 presents an analytic expression for the Rényi divergence term in Eq. (1.7), mitigating the extra effort that would arise relative to a SELBO calculation. The RELBO expression requires a double minimization over the encoder $V(y|x)$ and prior $q(y)$ that, for the prior, is not guaranteed to minimize at the true joint distribution $P(x,y)$. In Appendix C, this failure is studied generically. In Appendix D, a Gaussian Model computationally explores the issue. Though, as is conventional in VAEs, the prior is set, versus optimized. Our conclusions are presented in Section 5.

## 2. Rényi MI expression

To construct a Rényi VAE, definitions of Rényi versions of mutual information (MI) $I_\alpha$ and divergence $D_\alpha$ in the joint data/latent space must be provided. Work by Csiszár [18] and Shayevitz [19] suggest two. The one that proves convenient for our purposes is:

$$I_\alpha(P_X, W_{Y|X}) \doteq \min_{Q_Y} \int dx\, p(x) D_\alpha [W(y|x) \| q(y)] \doteq \min_{Q_Y} D_\alpha [W_{Y|X} \| P_Y | P_X] \tag{2.1}$$

with a Rényi divergence

$$D_\alpha\left[W(y|x)\|q(y)\right] \doteq \frac{1}{\alpha-1}\log\int dy W^\alpha(y|x) q^{1-\alpha}(y); \quad (\alpha > 0, \ \alpha \neq 1). \tag{2.2}$$

Shayevitz suggested writing the Rényi Mutual Information in Eq. (2.1) as, for $0 < \alpha < 1$,

$$I_\alpha(P_X, W_{Y|X}) = \min_{V_{Y|X}} \min_{Q_Y} \int dx\, p(x)\left[\frac{\alpha}{1-\alpha} D[V(y|x)\|W(y|x)] + D[V(y|x)\|q(y)]\right]. \tag{2.3}$$

(For $1 < \alpha < \infty$ the above is a max over $V_{Y|X}$). This form follows from the general expression in Appendix B.

Where does the min $V_{Y|X}$ occur? To see, make Eq. (2.3) explicit:

$$I_\alpha(P_X, W_{Y|X}) \doteq \min_{V_{Y|X}} \min_{Q_Y} f(V_{Y|X}, W_{Y|X}, Q_Y)$$

$$f(V_{Y|X}, W_{Y|X}, Q_Y) \doteq \int dx\, p(x) \int dy\, V(y|x) \log\left[\frac{V(y|x)}{q(y)}\right] \tag{2.4}$$

$$+ \frac{\alpha}{1-\alpha} \int dx\, p(x) \int dy\, V(y|x) \log\left[\frac{V(y|x)}{W(y|x)}\right].$$

Noting that $\alpha \log W(y|x) + (1-\alpha) q(y) = \log W^\alpha(y|x) q^{1-\alpha}(y)$, rewrite $f(V_{Y|X}, W_{Y|X}, Q_Y)$ as

$$f(V_{Y|X}, W_{Y|X}, Q_Y) = \frac{1}{1-\alpha} \int dx\, p(x) \int dy\, V(y|x) \log\left[\frac{V(y|x)}{q^{1-\alpha}(y) W^\alpha(y|x)}\right]. \tag{2.5}$$

Then consider

$$\log\left[\frac{V(y|x)}{q^{1-\alpha}(y) W^\alpha(y|x)}\right] = \log\left[\frac{V(y|x)}{V^*(y|x)}\right] + \log\left[\frac{V^*(y|x)}{q^{1-\alpha}(y) W^\alpha(y|x)}\right] \tag{2.6}$$

where

$$V^*(y|x) \doteq c_\alpha(x) q^{1-\alpha}(y) W^\alpha(y|x); \quad c_\alpha(x) \doteq \left[\int dy\, q^{1-\alpha}(y) W^\alpha(y|x)\right]^{-1}. \tag{2.7}$$

Thus

$$f(V_{Y|X}, W_{Y|X}, Q_Y) = \frac{1}{1-\alpha} \int dx\, p(x) \int dy\, V(y|x) \log\left[\frac{V(y|x)}{V^*(y|x)}\right]$$

$$+ \frac{1}{1-\alpha} \int dx\, p(x) \int dy\, V(y|x) \log\left[\frac{V^*(y|x)}{q^{1-\alpha}(y) W^\alpha(y|x)}\right]. \tag{2.8}$$

Then, using the definition in Eq. (2.7), write

$$f(V_{Y|X}, W_{Y|X}, Q_Y)$$

$$= \frac{1}{1-\alpha} \int dx\, p(x) D[V(y|x)\|V^*(y|x)] + \frac{1}{1-\alpha} \int dx\, p(x) \int dy\, V(y|x) \log c_\alpha(x) \tag{2.9}$$

$$= \frac{1}{1-\alpha} \int dx\, p(x) D[V(y|x)\|V^*(y|x)] + \frac{1}{1-\alpha} \int dx\, p(x) \log c_\alpha(x).$$

For $0 < \alpha < 1$ the first term is $\geq 0$ with equality for $V \to V^*$. Thus

$$\min_{V_{Y|X}} f(V_{Y|X}, W_{Y|X}, Q_Y) = f(V^*_{Y|X}, W_{Y|X}, Q_Y) \qquad (2.10)$$

with $V^*(y|x)$ given in Eq. (2.7).

In this form it is clear that the $\min V_{Y|X}$ is known, but $\min Q_Y$ is not necessarily $P_Y$ because $c_\alpha(x)$ depends on $Q_Y$ for $\alpha \neq 1$. For $\alpha = 1$, $c_{\alpha \to 1}(x) \doteq \left[\int dy W(y|x)\right]^{-1} = 1 (x \in \mathcal{X}_X)$ independent of $Q_Y$. In Appendix C, we provide a version of $I_\alpha(P_X, W_{Y|X})$ that shows this feature explicitly, and give an example of a distribution where $\min Q_Y \neq P_Y$. And also show that for the Shannon MI expression $\min Q_Y = P_Y$. Appendix D explores the issue for a Gaussian model.

In spite of this contrast between Rényi and Shannon it is still the case that for VAE purposes the prior $Q_Y$ is prescribed, mitigating this difference.

## 3. Rényi VAE

In the following assert $0 < \alpha < 1$ to not burden notation. Consider $I_\alpha(P_X, W_{Y|X})(x)$, for the each $x \in \mathcal{X}_X$ version of Eq. (2.3)

$$\begin{aligned} I_\alpha(P_X, W_{Y|X})(x) &\doteq \min_{V_{Y|X}} \min_{Q_Y} \left\{ D[V(y|x) \| q(y)] + \frac{\alpha}{1-\alpha} D[V(y|x) \| W(y|x)] \right\} \\ &= \min_{V_{Y|X}} \min_{Q_Y} \left\{ \int dy V(y|x) \log\left[\frac{V(y|x)}{q(y)}\right] + \frac{\alpha}{1-\alpha} \int dy V(y|x) \log\left[\frac{V(y|x)}{W(y|x)}\right] \right\}. \end{aligned} \qquad (3.1)$$

Use the Bayes-Laplace relation in the second Shannon divergence term in Eq. (3.1) to introduce a decoder $\tilde{V}(x|y) \doteq W(y|x) p(x)/q(y)$ and manipulate it to a VAE-like expression:

$$\begin{aligned} -D[V(y|x) \| W(y|x)] &= \mathbb{E}_{y \sim V(y|x)} \log \tilde{V}(y|x) + \int dy V(y|x) \log\left[\frac{q(y)}{V(y|x)}\right] - \log p(x) \\ &= \mathbb{E}_{y \sim V(y|x)} \log \tilde{V}(x|y) - D[V(y|x) \| q(y)] - \log p(x). \end{aligned} \qquad (3.2)$$

The use of Eq. (3.2) in Eq (3.1) provides one expression for $I_\alpha(P_X, W_{Y|X})(x)$. Eq (2.9) above provides another expression for $I_\alpha(P_X, W_{Y|X})(x)$. Equating the two to eliminate $I_\alpha(P_X, W_{Y|X})(x)$, and rearranging the result, provides the key result of this section:

$$\begin{aligned} &\log p(x) - D[V(y|x) \| V^*(y|x)] \\ &= \mathbb{E}_{y \sim V(y|x)} \log \tilde{V}(x|y) - \frac{1}{\alpha} D[V(y|x) \| q(y)] + \frac{1-\alpha}{\alpha} D_\alpha[W(y|x) \| q(y)] \end{aligned} \qquad (3.3)$$

Because $D[V(y|x) \| V^*(y|x)] \geq 0$, dropping it in Eq. (3.3) provides our RELBO expression

$$\log p(x) \geq \mathbb{E}_{y \sim V(y|x)} \log \tilde{V}(x|y) - \frac{1}{\alpha} D[V(y|x) \| q(y)] + \frac{1-\alpha}{\alpha} D_\alpha [W(y|x) \| q(y)] \quad (3.4)$$

It consists of a reconstruction term and an $\alpha$-dependent regularizer that would correspond to a $\beta$-VAE in the absence of the third term, a 'Rényi regularizer', that incorporates the true conditional $W(y|x)$. Thus, the RELBO has a different character than SELBO.

In the following section we evaluate the Rényi divergence in Eq.(3.4) using Gaussian models for both $W(y|x)$ and $q(y)$.

## 4. Evaluation of $D_\alpha[W(y|x) \| q(y)]$

The last, Rényi divergence, term in Eq.(3.4) incorporates the true posterior $W(y|x)$ that is viewed in the VAE world as intractable.[3,4] Here, we introduce a model to evaluate it, as it is required in the RELBO expression. For convenience, evaluate $D_{1-\alpha}[q(y) \| W(y|x)]$ noting the "skew symmetry" relation [15]

$$\frac{1-\alpha}{\alpha} D_\alpha[W(y|x) \| q(y)] = D_{1-\alpha}[q(y) \| W(y|x)] \quad (0 < \alpha < 1). \quad (4.1)$$

For current purposes, the exclusion $\alpha = 0$ is appropriate and the exclusion $\alpha = 1$ is irrelevant as it eliminates the Rényi regularizer term. $D_{1-\alpha}[q(y) \| W(y|x)]$ is a non-increasing function of $\alpha$ for $0 < \alpha < 1$, with scale set by the difference between its arguments. For notational convenience we evaluate $D_\alpha[q(y) \| W(y|x)]$ and then set $\alpha \to 1 - \alpha$.

In VAEs the prior is modelled as Gaussian, often using $q(y) \sim \mathcal{N}_y(0, I_y)$.[3,4]. Thus a Gaussian form for $W(y|x)$ would be convenient, as an analytic expression for $D_\alpha[P_i \| P_j]$ with both $P$'s Gaussian is known. A Gaussian model for $W(y|x)$ is available; namely, the probabilistic principal component analysis (P-PCA) model introduced by Roweis.[16], [17] It is a generalization of PCA that is used for data reduction/representation in a wide variety of fields.[20] The Roweis model connects data and latent space via

$$x(N_x) = C^R(N_x \times N_y) \cdot y(N_y) + \upsilon(N_x);$$
$$y \sim \prod_{k=1}^{N_y} \mathcal{N}_{y_k}(0,1), \quad \upsilon \sim \prod_{k=1}^{N_x} \mathcal{N}_{x_k}(0, \sigma^2), \quad \langle \upsilon \upsilon \rangle = \Sigma_x = \sigma^2 I_x. \quad (4.2)$$

$C^R(N_x, N_y)$ is the Roweis covariance matrix, obtained from an expectation over the data, and $y$ and $\upsilon$ are multidimensional i.i.d. Gaussians. The expected values of the latent variables depend on the additive noise with covariance matrix $\Sigma_x = \sigma^2 I_x$. The use of i.i.d. unit variance latents is well suited to VAEs as this is the conventional VAE choice of prior.[3,4] The required posterior in this model is also Gaussian:

$$W(y|x) = \mathcal{N}_y(\beta x, I_y - \beta C^R); \quad \beta \doteq C^{R,T}\left(C^R \cdot C^{R,T} + \Sigma_x\right)^{-1}. \tag{4.3}$$

The Rényi divergence of Gaussian distributions $P_i$ and $P_j$ is [21]

$$D_\alpha[P_i \| P_j] = \frac{1}{2}\alpha(\mu_i - \mu_j)^T \cdot (\Sigma_\alpha^*)^{-1} \cdot (\mu_i - \mu_j) - \frac{1}{2(\alpha-1)}\log\left[\frac{|\Sigma_\alpha^*|}{|\Sigma_i|^{1-\alpha}|\Sigma_j|^\alpha}\right] \tag{4.4}$$

with $\Sigma_\alpha^* \doteq (1-\alpha)\Sigma_i + \alpha\Sigma_j$.

The $\mu_i, \mu_j$ are the means and the $\Sigma_i, \Sigma_j$ the variances of the corresponding distributions $P_i, P_j$. We shall refer to the first term in Eq. (4.4) as the *scalar* term and the second as the *log det* term.

The 'map' to $D_\alpha[q(y)\|W(y|x)]$ associates $P_i \leftarrow q(y)$ and $P_j \leftarrow W(y|x)$. Thus, from Eqs. (4.2-4)

$$\mu_i = 0; \quad \Sigma_i = \sigma_i^2 I_y \to I_y;$$
$$\mu_j = \beta x; \quad \Sigma_j = I_y - \beta C^R. \tag{4.5}$$

We have set $\sigma_i^2 = 1$ for the standard prior. It is useful to make a dimensionless version $\Lambda \doteq C^R/\sigma$ of $C^R$. From Eqs. (4.3) and (4.5)

$$\Sigma_j^{-1} = \left[I_y - \beta C^R\right]^{-1} = \left[I_y - (C^R)^T\left[C^R(C^R)^T + \sigma^2 I_x\right]^{-1} C^R\right]^{-1}$$
$$\to \left[I_y - \Lambda^T\left[\Lambda\Lambda^T + I_x\right]^{-1}\Lambda\right]^{-1}; \quad \Lambda \doteq C^R/\sigma. \tag{4.6}$$

Note that $\Lambda(N_x \times N_y)$ and $\Lambda^T(N_y \times N_x)$.

To eliminate inversion of $[\Lambda\Lambda^T + I_x](N_x \times N_x)$ use the Woodbury matrix identity [22]

$$A^{-1} - A^{-1}U\left(C^{-1} + VA^{-1}U\right)^{-1}VA^{-1} = (A + UCV)^{-1} \tag{4.7}$$

and set $A = I_y; C = I_x; V = \Lambda; U = \Lambda^T$ to obtain

$$I_y - \Lambda^T\left(I_x + \Lambda\Lambda^T\right)^{-1}\Lambda = \left(I_y + \Lambda^T\Lambda\right)^{-1}. \tag{4.8}$$

Eq. (4.6) becomes

$$\Sigma_j^{-1} = \left[I_y - \beta C^R\right]^{-1} = I_y + \Lambda^T\Lambda. \tag{4.9}$$

This has the virtue of introducing $\Lambda^T\Lambda(N_y \times N_y)$ in the reduced, latent space. Use Eqs. (4.5) and (4.9) in Eq. (4.4) for $\Sigma_\alpha^*$:

$$\Sigma_\alpha^* \doteq (1-\alpha)\Sigma_i + \alpha\Sigma_j = (1-\alpha)I_y + \alpha(I_y - \beta C^R) = (1-\alpha)I_y + \alpha\left(I_y + \Lambda^T\Lambda\right)^{-1}. \tag{4.10}$$

Now use Woodbury in the form:

$$(A + B)^{-1} = A^{-1} - A^{-1}\left(AB^{-1} + I\right)^{-1} = A^{-1}\left[I - \left(AB^{-1} + I\right)^{-1}\right] \tag{4.11}$$

with $A = (1-\alpha)I_y$ and $B = \alpha\Sigma_j = \alpha(I_y + \Lambda^T\Lambda)^{-1}$ to obtain

$$(\Sigma_\alpha^*)^{-1} = \left\{\left[(1-\alpha)I_y + \alpha(I_y + \Lambda^T\Lambda)^{-1}\right]\right\}^{-1} = \frac{1}{1-\alpha}I_y - \frac{1}{1-\alpha}I_y\left[(1-\alpha)I_y\left(\frac{1}{\alpha}\right)(I_y + \Lambda^T\Lambda) + I_y\right]^{-1}$$

$$= \frac{1}{1-\alpha}\left\{I_y - \alpha\left[I_y + (1-\alpha)\Lambda^T\Lambda\right]^{-1}\right\}.$$

(4.12)

Also needed is the $\beta$ in Eq. (4.3):

$$\beta \doteq (C^R)^T\left[C^R(C^R)^T + \sigma^2 I_x\right]^{-1} \to \frac{1}{\sigma}\Lambda^T\left[I_x + \Lambda\Lambda^T\right]^{-1}; \quad \Lambda \doteq C^R/\sigma$$

$$\beta x \to \Lambda^T\left[I_x + \Lambda\Lambda^T\right]^{-1}\bar{x}; \quad \bar{x} \doteq x/\sigma$$

(4.13)

and

$$(\beta x)^T = x^T\beta^T \to \bar{x}^T\left\{\Lambda^T\left[I_x + \Lambda\Lambda^T\right]^{-1}\right\}^T$$

$$= \bar{x}^T\left\{\left[I_x + \Lambda\Lambda^T\right]^{-1}\right\}^T\Lambda = \bar{x}^T\left\{\left[I_x + \Lambda\Lambda^T\right]^{-1}\right\}\Lambda.$$

(4.14)

The last equality is true for any symmetric matrix such as $\Lambda\Lambda^T$.
Now use Eq. (4.7) in the direction

$$W = (A + UCV)^{-1} = A^{-1} - A^{-1}U(C^{-1} + VA^{-1}U)^{-1}VA^{-1}$$

(4.15)

with: $A = I_x; C = I_z; U = \Lambda; V = \Lambda^T$ to write

$$\Lambda^T(I_x + \Lambda\Lambda^T)^{-1} = \Lambda^T\left[I_x - \Lambda(I_z + \Lambda^T\Lambda)^{-1}\Lambda^T\right]$$

(4.16)

and similarly for the transpose term in Eq. (4.14). Inserting these in the scalar term of Eq. (4.4) produces, with use of Eq. (4.12)

$$(\beta x)^T(\Sigma_\alpha^*)^{-1}(\beta x) = \bar{x}^T S^\alpha \bar{x} = \bar{x}^T [1][2][3]\bar{x} \text{ with}$$

$$[1] = [3] \doteq \left[I_x - \Lambda(I_y + \Lambda^T\Lambda)^{-1}\Lambda^T\right]; \quad [2] \doteq \Lambda\frac{1}{1-\alpha}\left\{I_y - \alpha\left[I_y + (1-\alpha)\Lambda^T\Lambda\right]^{-1}\right\}\Lambda^T$$

(4.17)

where we have scaled the data: $\bar{x} \doteq x/\sigma$ with the latent standard deviation defined in Eq. (4.2). All inverses in Eq. (4.17) are expressed in the latent space via $\Lambda^T\Lambda$.

The above scalar term is formal. Use of Singular Value Decomposition (SVD) [22] will provide a convenient explicit expression. The SVD representation of $\Lambda \doteq C^R/\sigma$ is $\Lambda = ULV^T$, where $U(N_x \times N_y)$ [$V(N_y \times N_y)$] contains the left [right] singular values and $L(N_y \times N_y)$ is a diagonal matrix of the $N_y$ descending-ordered eigenvalues $\lambda_l$ $(l = 1, 2, ..., N_y)$ of the scaled Roweis covariance matrix. To use this SVD in the various terms in $S^\alpha$ of Eq.(4.17), first note that

$$\Lambda \doteq ULV^T; \quad \Lambda^T = \left(ULV^T\right)^T = VLU^T; \quad \Lambda^T\Lambda = VL^2V^T \tag{4.18}$$

Thus

$$\begin{aligned}\left(I_y + \Lambda^T\Lambda\right)^{-1} &= \left(I_y + VL^2V^T\right)^{-1} = \left[V\left(I_y + L^2\right)V^T\right]^{-1} \\ &= \left(V^T\right)^{-1}\left(I_y + L^2\right)^{-1}V^{-1} = V\left(I_y + L^2\right)^{-1}V^{-1}\end{aligned} \tag{4.19}$$

and

$$\Lambda\left(I_y + \Lambda^T\Lambda\right)^{-1}\Lambda^T = ULV^T\left[V\left(I_y + L^2\right)^{-1}V^{-1}\right]VLU^T = UL\left(I_y + L^2\right)^{-1}LU^T \tag{4.20}$$

where we have used the SVD properties:

$$U^TU = V^TV = VV^T = I_y \text{ and } V^{-1} = V^T \text{ and } V = \left(V^{-1}\right)^{-1} = \left(V^T\right)^{-1}. \tag{4.21}$$

Then, recognizing that $\left(I_y + L^2\right)^{-1}$ is a diagonal matrix

$$\Lambda\left(I_y + \Lambda^T\Lambda\right)^{-1}\Lambda^T = UG_d^{-1}U^T; \quad G_d^{-1} \doteq L\left(I_y + L^2\right)^{-1}L \tag{4.22.a}$$

with the second equality defining a diagonal matrix $G_d^{-1}$ with elements

$$\left(G_d^{-1}\right)_{kl} \doteq \delta_{kl}\lambda_l^2\left(1 + \lambda_l^2\right)^{-1}. \tag{4.22.b}$$

Eq. (4.22) provides [1] and [3] for Eq. (4.17):

$$[1] = [3] = I_x - \Lambda\left(I_y + \Lambda^T\Lambda\right)^{-1}\Lambda^T = I_x - UG_d^{-1}U^T. \tag{4.23}$$

For the middle term, [2], in Eq. (4.17) first work out

$$\begin{aligned}\left[I_y + (1-\alpha)\Lambda^T\Lambda\right]^{-1} &= \left[I_y + (1-\alpha)V^TL^2V\right]^{-1} = \left[V^T\left(I_y + (1-\alpha)L^2\right)V\right]^{-1} \\ &= V^T\left[\left(I_y + (1-\alpha)L^2\right)\right]^{-1}V.\end{aligned} \tag{4.24}$$

Thus

$$\begin{aligned}[2] &= \Lambda\left[\frac{1}{1-\alpha}\left\{I_y - \alpha\left[I_y + (1-\alpha)\Lambda^T\Lambda\right]^{-1}\right\}\right]\Lambda^T \\ &= \Lambda\left[\frac{1}{1-\alpha}\left\{V\left[I_y - \alpha\left[I_y + (1-\alpha)L^2\right]^{-1}\right]V^T\right\}\right]\Lambda^T \\ &= ULV^T\left[\frac{1}{1-\alpha}\left\{V\left[I_y - \alpha\left[I_y + (1-\alpha)L^2\right]^{-1}\right]V^T\right\}\right]VLU^T \\ &= UL\left[\frac{1}{1-\alpha}\left\{\left[I_y - \alpha\left[I_y + (1-\alpha)L^2\right]^{-1}\right]\right\}\right]LU^T.\end{aligned} \tag{4.25}$$

Define the diagonal matrix $G_d^\alpha$ as

$$G_d^\alpha \doteq L\left[\frac{1}{1-\alpha}\left\{\left[I_y - \alpha\left[I_y + (1-\alpha)L^2\right]^{-1}\right]\right\}\right]L \tag{4.26}$$

with elements

$$\left(G_d^\alpha\right)_{kl} \doteq \frac{1}{1-\alpha}\delta_{kl}\left\{\lambda_l\left[1 - \alpha\left[1 + (1-\alpha)\lambda_l^2\right]^{-1}\right]\lambda_l\right\}. \tag{4.27}$$

Note that

$$1 - \alpha\left[1 + (1-\alpha)\lambda_l^2\right]^{-1} = \frac{1 + (1-\alpha)\lambda_l^2 - \alpha}{1 + (1-\alpha)\lambda_l^2} = (1-\alpha)\left[\frac{1 + \lambda_l^2}{1 + (1-\alpha)\lambda_l^2}\right]. \tag{4.28}$$

Then write $G_d^\alpha$ in the manifestly positive form $(0 < \alpha < 1)$

$$\left(G_d^\alpha\right)_{kl} = \delta_{kl}\left[\frac{1 + \lambda_l^2}{1 + (1-\alpha)\lambda_l^2}\right]\lambda_l^2 = \delta_{kl}\frac{(1+\lambda_l^2)\lambda_l^2}{1 + \lambda_l^2 - \alpha\lambda_l^2} = \delta_{kl}\left[\frac{\lambda_l^2}{1 - \alpha\lambda_l^2/(1+\lambda_l^2)}\right]. \tag{4.29}$$

From Eq. (4.22.b)

$$\left(I_y - G_d^{-1}\right)_{kl} = \delta_{kl} - \delta_{kl}\frac{\lambda_l^2}{1+\lambda_l^2} = \delta_{kl}\frac{1}{1+\lambda_l^2}. \tag{4.30}$$

Combine Eq. (4.30) with Eq. (4.29) to write Eq. (4.17), using the SVD column orthogonality property of $U$ that $U^T U = I_y$, to define

$$S^\alpha \doteq \left[(I_x - UG_d^- U^T)U\right]G_d^\alpha\left[U^T(I_x - UG_d^- U^T)\right] = \left[U - UG_d^- U^T U\right]G_d^\alpha\left[U^T - U^T UG_d^- U^T\right]$$
$$= \left[U - UG_d^- I_y\right]G_d^\alpha\left[U^T - I_y G_d^- U^T\right] = U\left(\left[I_y - G_d^-\right]G_d^\alpha\left[I_y - G_d^-\right]\right)U^T. \tag{4.31}$$

Finally, define the diagonal matrix

$$H_d^\alpha \doteq \left[I_y - G_d^-\right]G_d^\alpha\left[I_y - G_d^-\right] \tag{4.32}$$

with elements

$$\left(H_d^\alpha\right)_{kl} = \delta_{kl}\left(\frac{1}{1+\lambda_l^2}\right)\left[\frac{\lambda_l^2}{1 - \alpha\lambda_l^2/(1+\lambda_l^2)}\right]\left(\frac{1}{1+\lambda_l^2}\right) = \delta_{kl}\frac{\lambda_l^2}{1+\lambda_l^2}\left[\frac{1}{(1+\lambda_l^2) - \alpha\lambda_l^2}\right] \tag{4.33}$$

so that Eq. (4.31) simplifies to $S^\alpha = UH_d^\alpha U^T$. Thus, the scalar divergence reduces to the expression

$$(\beta x)^T\left(\Sigma_\alpha^*\right)^{-1}(\beta x) = \bar{x}^T S^\alpha \bar{x} = \bar{x}^T UH_d^\alpha U^T \bar{x}. \tag{4.34}$$

It is readily evaluated from the SVD of $\Lambda = C^R/\sigma = ULV^T$ without requirement of matrix inversion. All that is required is the left singular value matrix $U\left(N_x \times N_y\right)$ and corresponding eigenvalues $\lambda_l \ (l=1,2,..,N_y)$ of $L$ that is available from the PCA of the data covariance matrix. It is input to the Rényi VAE for some value of $\alpha$.

Before evaluating the log det term in Eq. (4.4), rewrite it to see that it is non-negative.

$$\log\left[\frac{|\Sigma_\alpha^*|}{|\Sigma_i|^{1-\alpha}|\Sigma_j|^\alpha}\right] = \log\left[\frac{|\Sigma_\alpha^*|}{(1-\alpha)|\Sigma_i|\alpha|\Sigma_j|}\right] = \log\left[\frac{(1-\alpha)|\Sigma_i| + \alpha|\Sigma_j|}{(1-\alpha)|\Sigma_i|\alpha|\Sigma_j|}\right]$$

$$= \log\left[\frac{1+\delta|\Sigma_j|/|\Sigma_i|}{\delta|\Sigma_j|/|\Sigma_i|}\right] > 0; \quad \delta \doteq \frac{\alpha}{1-\alpha} \quad (0<\alpha<1)$$

(4.35)

To evaluate Eq. (4.35) the requisite explicit terms are

$$|I_y + \Lambda^T\Lambda| = |I_y + VL^2V^T| = |V(I_y + L^2)V^T|$$

$$= |V||I_y + L^2||V^T| = (\pm 1)^2|I_y + L^2| = \prod_{l=1}^{N_y}(1+\lambda_l^2)$$

(4.36)

using $|V| = |V^T| = \pm 1$ for orthonormal matrices. From Eq. (4.9) and Eq. (4.36), and noting that $\left|(I_y + \Lambda^T\Lambda)^{-1}\right| = |I_y + \Lambda^T\Lambda|^{-1}$, provides

$$|\Sigma_j|/|\Sigma_i| = |I_y - \beta C^R|/|I_y| = |(I_y + \Lambda^T\Lambda)^{-1}|/|I_y| = \left[\prod_{l=1}^{N_y}(1+\lambda_l^2)\right]^{-1}$$

(4.37)

Thus Eq. (4.35) with use of Eq. (4.37) is simply

$$\log\left[\frac{|\Sigma_\alpha^*|}{|\Sigma_i|^{1-\alpha}|\Sigma_j|^\alpha}\right] = \log\left[\frac{1+\delta\prod_{l=1}^{N_y}\left[(1+\lambda_l^2)^{-1}\right]}{\delta\prod_{l=1}^{N_y}\left[(1+\lambda_l^2)^{-1}\right]}\right] = \log\left[\frac{(1-\alpha)+\alpha\prod_{l=1}^{N_y}\left[(1+\lambda_l^2)^{-1}\right]}{\alpha\prod_{l=1}^{N_y}\left[(1+\lambda_l^2)^{-1}\right]}\right]$$

(4.38)

Both the log det term in Eq. (4.38) and scalar term in Eq. (4.34) that are the ingredients of Eq. (4.4) as applied to $D_\alpha[q(y)\|W(y|x)]$ are expressible by the eigenvalues and left eigenvectors that are readily obtained from the SVD of the data covariance matrix.

Finally, from Eq. (4.1), set $\alpha \to 1-\alpha$ in Eq. (3.4) for the Rényi regularizer term:

$$D_{1-\alpha}[q(y)\|W(y|x)] = \log\left[\frac{\alpha+(1-\alpha)\prod_{l=1}^{N_y}\left[(1+\lambda_l^2)^{-1}\right]}{(1-\alpha)\prod_{l=1}^{N_y}\left[(1+\lambda_l^2)^{-1}\right]}\right] = \log\left[\frac{\alpha}{1-\alpha}\prod_{l=1}^{N_y}\left[(1+\lambda_l^2)\right]+1\right]$$ (4.39)

## 5. Conclusion

Variational autoencoders' disentanglement capabilities are improved with a multiplicative factor, $\beta$, on the regularizer. The Shannon-based-entropy ELBO is correct only for $\beta=1$. Use of Rényi entropy does provide a VAE with a multiplicative factor to balance regularizer and reconstruction terms. In addition, the RELBO incorporates a new Rényi divergence term that requires the true conditional, $W(y|x)$. As shown in Section 4, it can be evaluated essentially analytically by using a Gaussian P-PCA model, consonant with the Gaussian character of VAEs.

That the minimizer to obtain the RELBO is not necessarily obtained for prior $q(y) \to p(y)$ is a difference with the Shannon ELBO. We explore the issue with discrete and Gaussian models. Nevertheless, as typically the prior is specified in ELBOs, the distinction is not significant.

**Appendix A.** Failure of the $\beta$ - ELBO expression.

Suppose we rewrite the exact Shannon VAE expression

$$\log p(x) - D[q(y|x) \| p(y|x)] = \mathbb{E}_{y \sim q(y|x)} \log p(x|y) - D[q(y|x) \| p(y)] \quad (A.1)$$

into the identity

$$\lambda \log p(x) - D[q(y|x) \| p(y|x)] = \{\lambda \mathbb{E}_{y \sim q(y|x)} \log p(x|y) - D[q(y|x) \| p(y)]\}$$
$$+ (1-\lambda)\{D[q(y|x) \| p(y)] - D[q(y|x) \| p(y|x)]\} \quad (A.2)$$

If the second {} term is dropped, Eq. (A.2) yields

$$\lambda \log p(x) - D[q(y|x) \| p(y|x)] = \{\lambda \mathbb{E}_{y \sim q(y|x)} \log p(x|y) - D[q(y|x) \| p(y)]\}. \quad (A.3)$$

With $\lambda \to 1/\beta$ Eq. (A.3) reads

$$\log p(x) - \beta D[q(y|x) \| p(y|x)] = \mathbb{E}_{y \sim q(y|x)} \log p(x|y) - \beta D[q(y|x) \| p(y)], \quad (A.4)$$

and the $\beta$ - ELBO expression follows:

$$\log p(x) \geq \mathbb{E}_{y \sim q(y|x)} \log p(x|y) - \beta D[q(y|x) \| p(y)]. \quad (A.5)$$

Can the second {} term in Eq.(A.2) be dropped to obtain a variational bound? It is the difference of two non-negative divergences

$$\{D[q(y|x) \| p(y)] - D[q(y|x) \| p(y|x)]\} = -\int dy\, q(y|x) \log\left[\frac{p(x,y)}{p(x)p(y)}\right] \quad (A.6)$$

and that difference has no definite sign. For $q(y|x) \to p(y|x)$ it becomes the mutual information $I(P_X; P_Y)$ that is non-negative unless the random variables $X$ and $Y$ are uncorrelated. Thus, there is no justification to drop this term to obtain a $\beta$ - ELBO.

**Appendix B.** A useful identity for $I_\alpha(P,W)$.

The governing equation for $I_\alpha(P,W)$ in Eqs.(2.1-2.3) arises from a useful identity [15] that relates Rényi and Shannon divergences. Generically it is:

$$\alpha D[R\|P] + (1-\alpha)D[R\|Q] = D[R\|p_\alpha] + (1-\alpha)D_\alpha[P\|Q] \tag{B.1}$$

where

$$p_\alpha \doteq P^\alpha Q^{1-\alpha} \Big/ \int P^\alpha Q^{1-\alpha} \tag{B.2}$$

is a mixed pdf.

Start from the left-hand side of Eq. (B.1)

$$\alpha D[R\|P] + (1-\alpha)D[R\|Q] = \alpha \int R\log\frac{R}{P} + (1-\alpha)\int R\log\frac{R}{Q}$$

$$= [\alpha + (1-\alpha)]\int R\log R - \int R\log P^\alpha - \int R\log Q^{1-\alpha} \tag{B.3}$$

$$= \int R\log R - \int R\log P^\alpha Q^{1-\alpha}.$$

Then write

$$\log[P^\alpha Q^{1-\alpha}] = \log\left[\frac{P^\alpha Q^{1-\alpha}}{\int P^\alpha Q^{1-\alpha}}\right] + \log\left[\int P^\alpha Q^{1-\alpha}\right] \tag{B.4}$$

$$\therefore \int R\log P^\alpha Q^{1-\alpha} = \int R\log p_\alpha + \int R\log\int P^\alpha Q^{1-\alpha}$$

where the mixed pdf is defined in Eq. (B.2). And,

$$\int R\log R - \int R\log P^\alpha Q^{1-\alpha} = \int R\log R - \int R\log p_\alpha - \int R\log\int P^\alpha Q^{1-\alpha}$$

$$= D[R\|p_\alpha] - \int R\log\int P^\alpha Q^{1-\alpha} = D[R\|p_\alpha] - (1)\log\int P^\alpha Q^{1-\alpha} \tag{B.5}$$

$$= D[R\|p_\alpha] + (1-\alpha)D_\alpha[P\|Q]$$

which is the right-hand side of Eq. (B.1).

With rearrangement and application to the conditionals of interest, Eq. (B.1) becomes

$$(1-\alpha)D_\alpha(W_{Y|X}\|Q_Y) = \alpha D(R_{Y|X}\|W_{Y|X}) + (1-\alpha)D(R_{Y|X}\|Q_Y) - D(R_{Y|X}\|P^\alpha_{Y|X}), \tag{B.6}$$

noting the definition $P^\alpha_{Y|X} \doteq V^*(y|x) = W^\alpha(y|x)q^{1-\alpha}(y)c(x)$ in Eq.(2.7). Then, set $R_{Y|X} \to P^\alpha_{Y|X}$ to get

$$(1-\alpha)D_\alpha(W_{Y|X}\|Q_Y) = \alpha D(P^\alpha_{Y|X}\|W_{Y|X}) + (1-\alpha)D(P^\alpha_{Y|X}\|Q_Y) \tag{B.7}$$

or

$$D_\alpha(W_{Y|X}\|Q_Y) = \min_{R_{Y|X}}\left\{\frac{\alpha}{1-\alpha}D(R_{Y|X}\|W_{Y|X}) + D(R_{Y|X}\|Q_Y)\right\}. \tag{B.8}$$

**Appendix C.** Double minimization of a dichotomic joint probability.

First, we show directly why $\min Q_Y$ is not necessarily obtained for $Q_Y \to P_Y$ in the Rényi case for the mutual information in Eq.(2.3). From Eq. (2.9)

$$f(V_{Y|X}, W_{Y|X}, Q_Y) = \frac{1}{1-\alpha}\left[\int dx p(x) D[V(y|x) \| V^*(y|x)] + \int dx p(x) \log c_\alpha(x)\right]. \quad (C.1)$$

Set $V(y|x) \to V^*(y|x)$ and write

$$f(V_{Y|X}, W_{Y|X}, Q_Y) = \min_{Q_Y}\left\{\frac{1}{1-\alpha}\int dx p(x) \log\left[\frac{c_\alpha(x)}{c_\alpha^*(x)}\right]\right\} + \frac{1}{1-\alpha}\int dx p(x) \log[c_\alpha^*(x)]$$

with (C.2)

$$c_\alpha^*(x) \doteq \left[\int dy W^\alpha(y|x) p^{1-\alpha}(y)\right]^{-1}; \quad c_\alpha(x) \doteq \left[\int dy W^\alpha(y|x) q^{1-\alpha}(y)\right]^{-1}.$$

If it is the case that

$$\min_{Q_Y}\left\{\frac{1}{1-\alpha}\int dx p(x) \log\left[\frac{c_\alpha(x)}{c_\alpha^*(x)}\right]\right\} \geq 0 \quad (C.3)$$

then $Q_Y \to P_Y$ would be a minimizer but Eq. (C.3) can be written as

$$\frac{1}{1-\alpha}\log\left[\frac{c_\alpha(x)}{c_\alpha^*(x)}\right] = D_\alpha[W(y|x) \| q(y)] - D_\alpha[W(y|x) \| p(y)] \quad (C.4)$$

and nothing says this minimizes at zero, where the two divergences are equal.

It would be of interest to introduce that as a constraint in the Rényi VAE.

To show that in fact the minimized value can be negative, write Eq. (C.3) using Eq. (C.4) and the definitions in Eq. (C.2) as

$$F_\alpha(Q_Y, P_Y) \doteq \frac{1}{1-\alpha}\int dx p(x) \log\left[\frac{c(x)}{c^*(x)}\right]$$

$$= -\frac{1}{1-\alpha}\int dx p(x) \log \int dy\left[R^\alpha(y|x)[s(y)]^{1-\alpha}\right]; \quad (C.5)$$

with

$$s(y) \doteq q(y)/p(y).$$

In writing Eq. (C.5) we defined

$$R^\alpha(y|x) \doteq \frac{p(y) W^\alpha(x|y)}{\int dy p(y) W^\alpha(x|y)} \quad \text{s.t.} \quad \int dy R^\alpha(y|x) = 1; \quad x \in \mathcal{X}_x \quad (C.6)$$

using Bayes to introduce the conditioned-on-$x$ conditional probability $R^\alpha(y|x)$. This is a convenient form with which to explore the sign of $F_\alpha(Q_Y, P_Y)$. In particular, consider a dichotomic joint probability; $\mathcal{X}_x, \mathcal{X}_y = \{0,1\}$. Making use of the constraint in Eq. (C.6), Eq.(C.5) evaluates to

$$F_\alpha = -\frac{1}{1-\alpha}\left\{\begin{array}{l}p_X(0)\log\left[s^{1-\alpha}(1)+R^\alpha(0|0)\left(s^{1-\alpha}(0)-s^{1-\alpha}(1)\right)\right]\\+p_X(1)\log\left[s^{1-\alpha}(1)+R^\alpha(1|0)\left(s^{1-\alpha}(0)-s^{1-\alpha}(1)\right)\right]\end{array}\right\}.$$

Noting that the $R^\alpha(y|x) < 1$, and supposing that $s(0) = q(0)/p(0) \gg 1$ and therefore $s(1) = q(1)/p(1) \ll 1$, then

$$F_\alpha \to -\frac{1}{1-\alpha}\left\{p_X(0)\log R^\alpha(0|0)s^{1-\alpha}(0) + p_X(1)\log R^\alpha(1|0)s^{1-\alpha}(0)\right\}$$

$$= -\frac{1}{1-\alpha}\left\{\log s^{1-\alpha}(0) + p_X(0)R^\alpha(0|0) + p_X(1)\log R^\alpha(1|0)\right\} \qquad (C.7)$$

$$\to -\frac{1}{1-\alpha}\log s^{1-\alpha}(0) = -\log s(0).$$

Therefore, with $s(0) \gg 1$: $I_\alpha < 0$ $(0 < \alpha < 1)$.

In contrast, for Shannon MI, the double opt is satisfied by $V(y|x) \to W(y|x);\ q(y) \to p(y)$. The normalization factor of the general Rényi expression in Eq. (2.7) for $\alpha \to 1$ is

$$V^*_{\alpha \to 1}(y|x) \doteq W(y|x)c_{\alpha=1}(x); \quad c_{\alpha=1}(x) \doteq \left[\int dy W(y|x)\right]^{-1} = 1 \quad (x \in \mathcal{X}_x)$$
$$\therefore V^*_{\alpha \to 1}(y|x) = W(y|x). \qquad (C.8)$$

Only for $\alpha \to 1$ is the optimized $V_\alpha(y|x)$ independent of the prior $q(y)$. The Shannon MI then becomes

$$\min_{V_{Y|X}} I_{\alpha=1}(P_X, V_{Y|X}) = I_{\alpha=1}(P_X, W_{Y|X}) \qquad (C.9)$$

with

$$I_{\alpha=1}(P_X, W_{Y|X}) = \min_{Q_Y} \int dx\, p(x) \int dy\, W(y|x) \log\left[\frac{W(y|x)}{q(y)}\right]$$

$$= \min_{Q_Y} \int dx\, p(x) \int dy\, W(y|x) \log\left\{\left[\frac{W(y|x)p(x)}{p(x)p(y)}\right]\left[\frac{p(y)}{q(y)}\right]\right\}$$

$$= \min_{Q_Y}\left(\int dx\, p(x) \int dy\, W(y|x) \log\left[\frac{W(y|x)p(x)}{p(x)p(y)}\right] + \int dx\, p(x) \int dy\, W(y|x) \log\left[\frac{p(y)}{q(y)}\right]\right)$$

$$= \min_{Q_Y}\left(\int dx\, p(x) \int dy\, W(y|x) \log\left[\frac{W(y|x)p(x)}{p(x)p(y)}\right] + \int dy \int dx\, W(y|x) p(x) \log\left[\frac{p(y)}{q(y)}\right]\right)$$

$$= \min_{Q_Y}\left(\int dx\, p(x) \int dy\, W(y|x) \log\left[\frac{W(y|x)p(x)}{p(x)p(y)}\right] + \int dy\, p(y) \log\left[\frac{p(y)}{q(y)}\right]\right) \qquad (C.10)$$

$$= I(P_X, W_{Y|X}) + \min_{q_y} D(P_Y \| Q_Y).$$

By Jensen's inequality $D(P_Y \| Q_Y) \geq 0$. Therefore, the Shannon MI is minimized with $V(y|x) \to W(y|x); \; q(y) \to p(y)$. In obtaining this result it was crucial that $V^*(y|x) = W(y|x)$. Then, $V(y|x)p(x) = W(y|x)p(x) \doteq p(x,y)$ and $\int dx p(x,y) = p(y)$.

**Appendix D** RELBO double minimization of a Gaussian model.

The RELBO expression in Eq. (3.1) is a double minimization over prior $q(y)$ and encoder $V(y|x)$ that is not guaranteed to minimize at the true joint distribution $P(x,y)$, in contrast with the ELBO that does minimize this way, as shown in Appendix C. To investigate the issue, express $I_\alpha(Q_Y, P_Y)$ as

$$I_\alpha(Q_Y, P_Y) = \bar{I}_\alpha(Q_Y, P_Y) + \frac{1}{1-\alpha} \int dx p(x) \log\left[c^*(x)\right]$$

$$c_\alpha^*(x) \doteq \left[\int dy W^\alpha(y|x) p^{1-\alpha}(y)\right]^{-1}; \quad c_\alpha(x) \doteq \left[\int dy W^\alpha(y|x) q^{1-\alpha}(y)\right]^{-1} \quad \text{(D.1)}$$

$$\bar{I}_\alpha(Q_Y, P_Y) \doteq \min_{Q_Y} \left\{ \frac{1}{1-\alpha} \int dx p(x) \log\left[\frac{c(x)}{c^*(x)}\right] \right\}$$

In this form, if $\min_{Q_Y} \bar{I}_\alpha(Q_Y, P_Y) = \bar{I}_\alpha(P_Y, P_Y) = 0$ then indeed $Q_Y \to P_Y$ is the minimizer. However, expressing $\bar{I}_\alpha(Q_Y, P_Y)$ in terms of Rényi divergences via

$$\frac{1}{1-\alpha} \int dx p(x) \log\left[\frac{c(x)}{c^*(x)}\right]$$

$$= \int dx p(x) \left[ D_\alpha(W(y|x) \| q(y))(x) - D_\alpha(W(y|x) \| p(y))(x) \right] \quad \text{(D.2)}$$

$$= D_\alpha(W(y|x) \| q(y) | p(x)) - D_\alpha(W(y|x) \| p(y) | p(x))$$

with the definitions

$$D_\alpha(W(y|x) \| q(y))(x) \doteq \frac{1}{\alpha-1} \int dy W^\alpha(y|x) q^{1-\alpha}(y)$$

$$D_\alpha(W(y|x) \| q(y) | p(x)) \doteq \int dx p(x) D_\alpha(W(y|x) \| q(y))(x) \quad \text{(D.3)}$$

shows that, as a difference of two non-negative quantities, $\bar{I}_\alpha(Q_Y, P_Y)$ could be negative for $Q_Y \not\to P_Y$.

In this Appendix, all plots are generated with Mathematica. [23]

Before examining a Gaussian model, consider the *independent* limit where $W(y|x) \to p(y)$. Then

$$\bar{I}_\alpha(Q_Y, P_Y) \to D_\alpha(p(y) \| q(y) | p(x)) - D_\alpha(p(y) \| p(y) | p(x))$$

$$= \int dx p(x) \left[ D_\alpha(p(y) \| q(y))(x) \right] = D_\alpha(p(y) \| q(y)) \geq 0; \quad (0 < \alpha < \infty) \quad \text{(D.4)}$$

Thus $\min_{Q_Y} \bar{I}_\alpha(Q_Y, P_Y) = \bar{I}_\alpha(P_Y, P_Y) = 0$ s.t. $Q_Y \to P_Y$ is the minimizer for all $\alpha$ values. Independent limit plots are displayed in Figure 1 for $\alpha < 1$ and $\alpha > 1$ showing the scale of $\bar{I}_\alpha(Q_Y, P_Y)$.

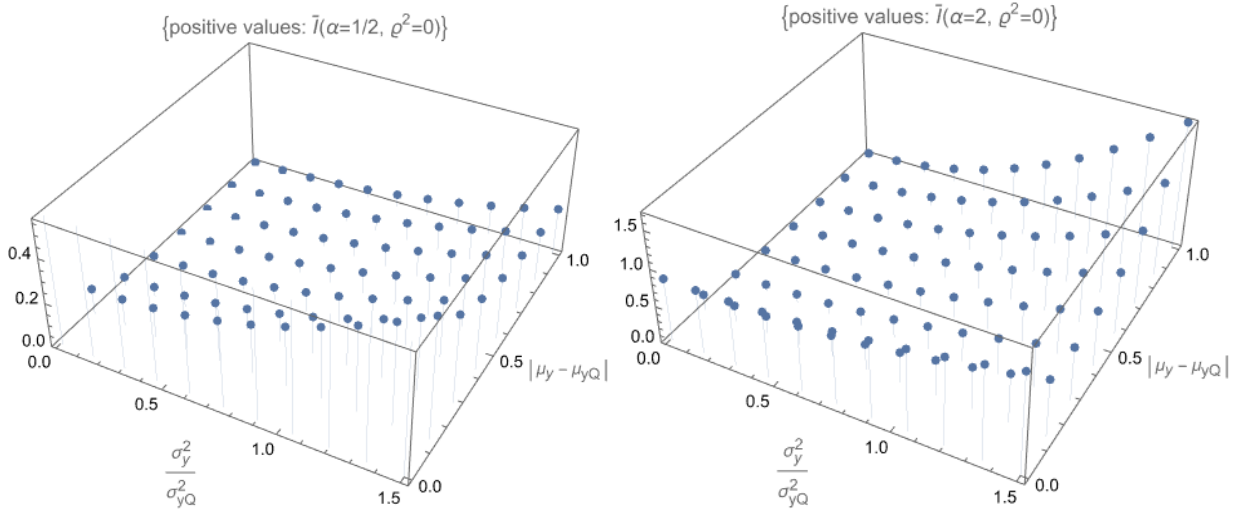

Figure 1. $\bar{I}_\alpha^{GM}(P_Y, Q_Y)$ for independent $W(y|x) \to p(y)$ for $\alpha < 1$ and $\alpha > 1$. All values are non-negative indication that $\min_{Q_Y} \to P_Y$.

For the more relevant correlated cases, introduce a Gaussian model (GM) that does have an explicit form for the desired Rényi divergence. The Gaussian model divergences, with $s(y) = \{q(y), p(y)\}$, are [21]

$$\frac{1}{\alpha-1}\log \int dy\, W^\alpha(y|x) s^{1-\alpha}(y) = D_\alpha\left(\mathcal{N}(\mu_0, \sigma_0^2) \| \mathcal{N}(\mu_1, \sigma_1^2)\right)$$

$$= \frac{\alpha(\mu_1 - \mu_0)^2}{2\sigma_\alpha^2} + \frac{1}{2(1-\alpha)}\log\left[\frac{\sigma_\alpha^2}{(\sigma_0^2)^{1-\alpha}(\sigma_1^2)^\alpha}\right] \quad \text{(D.5)}$$

$$\sigma_\alpha^2 \doteq (1-\alpha)\sigma_0^2 + \alpha\sigma_1^2.$$

The required conditional distribution is [17]

$$W(y|x) = \mathcal{N}(y|\mu_C(x), \sigma_C^2);$$

$$\mu_C(x) \doteq \mu_y + \frac{\sigma_y}{\sigma_x}\rho(x-\mu_x); \quad \sigma_C^2 \doteq (1-\rho^2)\sigma_y^2 \tag{D.6}$$

and provides the parameter associations:

$$\begin{aligned}\mu_0, \sigma_0^2 &\to W(y|x) \to \mu_C(x), \sigma_C^2 \\ \mu_1, \sigma_1^2 &\to q_y \to \mu_{y,Q}, \sigma_{y,Q}^2 \\ \mu_1, \sigma_1^2 &\to p_y \to \mu_y, \sigma_y^2.\end{aligned} \tag{D.7}$$

There is a Gaussian model constraint that the combination variance must be non-negative, $\sigma_\alpha^2 \geq 0$. The desired GM divergence difference is

$$D_\alpha\big(\mathcal{N}(\mu_C, \sigma_C^2) \| \mathcal{N}(\mu_{y,Q}, \sigma_{y,Q}^2)\big)(x) - D_\alpha\big(\mathcal{N}(\mu_C, \sigma_C^2) \| \mathcal{N}(\mu_y, \sigma_y^2)\big)(x)$$

$$= \frac{\alpha}{2}\left[\frac{(\mu_C(x)-\mu_{y,Q})^2}{(1-\alpha)\sigma_C^2 + \alpha\sigma_{y,Q}^2} - \frac{(\mu_C(x)-\mu_y)^2}{(1-\alpha)\sigma_C^2 + \alpha\sigma_y^2}\right] \tag{D.8}$$

$$+ \frac{1}{2(1-\alpha)}\left\{\log\left[\frac{(1-\alpha)\sigma_C^2 + \alpha\sigma_{y,Q}^2}{(\sigma_C^2)^{1-\alpha}(\sigma_{y,Q}^2)^\alpha}\right] - \log\left[\frac{(1-\alpha)\sigma_C^2 + \alpha\sigma_y^2}{(\sigma_C^2)^{1-\alpha}(\sigma_y^2)^\alpha}\right]\right\}$$

where one needs to check that both

$$\sigma_{\alpha,y}^2 \doteq (1-\alpha)\sigma_C^2 + \alpha\sigma_y^2 \geq 0; \quad \sigma_{\alpha,y,Q}^2 \doteq (1-\alpha)\sigma_C^2 + \alpha\sigma_{y,Q}^2 \geq 0. \tag{D.9}$$

It is convenient to write the second, log term in the above difference as

$$\frac{1}{2(1-\alpha)}\left\{\log\left[\frac{(1-\alpha)\sigma_C^2 + \alpha\sigma_{y,Q}^2}{(\sigma_C^2)^{1-\alpha}(\sigma_{y,Q}^2)^\alpha}\right] - \log\left[\frac{(1-\alpha)\sigma_C^2 + \alpha\sigma_y^2}{(\sigma_C^2)^{1-\alpha}(\sigma_y^2)^\alpha}\right]\right\}$$

$$= \frac{1}{2(1-\alpha)}\left\{\log\left[\frac{(1-\alpha)\sigma_C^2 + \alpha\sigma_{y,Q}^2}{(1-\alpha)\sigma_C^2 + \alpha\sigma_y^2}\right]\left[\frac{(\sigma_y^2)^\alpha}{(\sigma_{y,Q}^2)^\alpha}\right]\right\}. \tag{D.10}$$

It is $x$-independent; so, this contribution to $\bar{I}_\alpha^{GM}(Q_Y, P_Y)$ in Eq. (D.10) is

$$_2\bar{I}_\alpha^{GM} \doteq \frac{1}{2(1-\alpha)}\left\{\log\left[\frac{(1-\alpha)\sigma_C^2 + \alpha\sigma_{y,Q}^2}{(1-\alpha)\sigma_C^2 + \alpha\sigma_y^2}\right]\left[\frac{(\sigma_y^2)^\alpha}{(\sigma_{y,Q}^2)^\alpha}\right]\right\}. \tag{D.11}$$

The first term in Eq. (D.10) is

$$_1\bar{I}_\alpha^{GM} \doteq \frac{1}{2(\alpha-1)}\int dx\, p(x)\left\{\frac{\alpha}{2}\left[\frac{(\mu_C(x)-\mu_{y,Q})^2}{(1-\alpha)\sigma_C^2 + \alpha\sigma_{y,Q}^2} - \frac{(\mu_C(x)-\mu_y)^2}{(1-\alpha)\sigma_C^2 + \alpha\sigma_y^2}\right]\right\}. \tag{D.12}$$

To carry out the $x$ integration it is convenient to define

$$\mu_C(x) \doteq c_1 + c_2 x; \text{ with } c_1 = \mu_y - \frac{\sigma_y}{\sigma_x}\rho\mu_x; \ c_2 = \frac{\sigma_y}{\sigma_x}\rho. \tag{D.13}$$

Then

$$_1\bar{I}_\alpha^{GM} = \frac{\alpha}{2}\int dx p(x)\left\{\frac{\alpha}{2}\left[\frac{\left[(c_1-\mu_{y,Q})+c_2 x\right]^2}{(1-\alpha)\sigma_C^2+\alpha\sigma_{y,Q}^2} - \frac{\left[(c_1-\mu_y)+c_2 x\right]^2}{(1-\alpha)\sigma_C^2+\alpha\sigma_y^2}\right]\right\}. \tag{D.14}$$

The $\mathcal{N}_x(x|\mu_x,\sigma_x^2)$ expectations are $\mathbb{E}_{\mathcal{N}_x}(1)=1$; $\mathbb{E}_{\mathcal{N}_x}(x)=\mu_x$; $\mathbb{E}_{\mathcal{N}_x}(x^2)=\mu_x^2+\sigma_x^2$.

These expectations on $\left[(c_1-\mu_{y,Q})+c_2 x\right]^2 = (c_1-\mu_{y,Q})^2 + 2(c_1-\mu_{y,Q})c_2 x + c_2^2 x^2$ lead to

$$\mathbb{E}_{\mathcal{N}_x}\left[(c_1-\mu_{y,Q})^2 + 2(c_1-\mu_{y,Q})c_2 x + c_2^2 x^2\right]$$
$$= (c_1-\mu_{y,Q})^2 + 2(c_1-\mu_{y,Q})c_2\mu_x + c_2^2(\mu_x^2+\sigma_x^2) \tag{D.15}$$
$$= \left[(c_1-\mu_{y,Q})+c_2\mu_x\right]^2 + c_2^2\sigma_x^2.$$

Use of $c_1=\mu_y-\frac{\sigma_y}{\sigma_x}\rho\mu_x$; $c_2=\frac{\sigma_y}{\sigma_x}\rho$ shows that $\left[(c_1-\mu_{y,Q})+c_2\mu_x\right]^2 = (\mu_y-\mu_{y,Q})^2$ and with $c_2^2\sigma_x^2 = \sigma_y^2\rho^2$ provides

$$_1\bar{I}_\alpha^{GM} \doteq \frac{\alpha}{2}\left[\frac{(\mu_y-\mu_{y,Q})^2+\sigma_y^2\rho^2}{(1-\alpha)\sigma_C^2+\alpha\sigma_{y,Q}^2} - \frac{\sigma_y^2\rho^2}{(1-\alpha)\sigma_C^2+\alpha\sigma_y^2}\right]. \tag{D.16}$$

Thus

$$\bar{I}_\alpha(Q_Y,P_Y) = \min_{Q_Y}\left\{\frac{1}{1-\alpha}\int dx p(x)\log\left[\frac{c(x)}{c^*(x)}\right]\right\} = \min_{Q_Y}\left\{_1\bar{I}_\alpha^{GM}(Q_Y,P_Y) +\ _2\bar{I}_\alpha^{GM}(Q_Y,P_Y)\right\} \tag{D.17}$$

with $_1\bar{I}_\alpha^{GM}(Q_Y,P_Y)$ and $_2\bar{I}_\alpha^{GM}(Q_Y,P_Y)$ given respectively in Eqs. (D.16) and (D.11). Putting these terms together and using $\sigma_C^2 \doteq (1-\rho^2)\sigma_y^2$, the explicit final form is:

$$\bar{I}_\alpha^{GM}(P_Y,Q_Y) = \frac{\alpha}{2}\left[\frac{(\mu_y-\mu_{y,Q})^2/\sigma_{y,Q}^2 + \rho^2(\sigma_y^2/\sigma_{y,Q}^2)}{(1-\alpha)(1-\rho^2)(\sigma_y^2/\sigma_{y,Q}^2)+\alpha} - \frac{\rho^2}{(1-\alpha)(1-\rho^2)+\alpha}\right]$$
$$+ \frac{1}{2(1-\alpha)}\log\left\{\left[\frac{(1-\alpha)(1-\rho^2)+\alpha/(\sigma_y^2/\sigma_{y,Q}^2)}{(1-\alpha)(1-\rho^2)+\alpha}\right]\left[\frac{\sigma_y^2}{\sigma_{y,Q}^2}\right]^\alpha\right\} \tag{D.18}$$

Before evaluating Eq. (D.18), it is worth examining the opposite to the independent limit: namely, the completely-correlated limit, $\rho \to 1$, where $\sigma_C^2 \doteq (1-\rho^2)\sigma_y^2 \to 0$. From Eq. (D.9)

$$\sigma_{\alpha,y}^2 \doteq (1-\alpha)\sigma_C^2 + \alpha\sigma_y^2 \to \alpha\sigma_y^2 \geq 0$$
$$\sigma_{\alpha,y,Q}^2 \doteq (1-\alpha)\sigma_C^2 + \alpha\sigma_{y,Q}^2 \to \alpha\sigma_{y,Q}^2 \geq 0$$
(D.19)

indicating that there is no integral restriction for any $\alpha$ value. For $\rho \to 1$, $W_{\rho \to 1}(y|x) \to \delta(y - \mu_C(x|\rho \to 1))$ and $\mu_C(x|\rho \to 1) = \mu_y + \frac{\sigma_y}{\sigma_x}(x - \mu_x)$, from Eq.(D.13). Thus $y(x) = \mu_C(x, \rho \to 1) = \mu_y + \frac{\sigma_y}{\sigma_x}(x - \mu_x)$. Then Eq. (D.18) for $\overline{I}_\alpha^{GM}(P_Y, Q_Y)$ for $\rho \to 1$, after some reshaping, takes the simple form

$$\overline{I}_\alpha^{GM}(P_Y, Q_Y | \rho \to 1) = \frac{1}{2}\left\{\frac{(\mu_y - \mu_{y,Q})^2}{\sigma_{y,Q}^2} + \left[\sigma_y^2/\sigma_{y,Q}^2 - 1 - \log(\sigma_y^2/\sigma_{y,Q}^2)\right]\right\}.$$
(D.20)

It is plotted in Figure 2. Note that it is $\alpha$ independent, as both standard deviations in Eq. (D.19) are proportional to $\alpha$. For $\sigma_y^2/\sigma_{y,Q}^2 \geq 1$, $\min Q_Y \to P_Y$ for all other parameter values. For $\sigma_y^2/\sigma_{y,Q}^2 \leq 1$, $\overline{I}_\alpha^{GM}(P_Y, Q_Y)$ can go negative. Though, as the difference of means $(\mu_y - \mu_{y,Q})^2$ increases, $\min Q_Y \to P_Y$ is more robust.

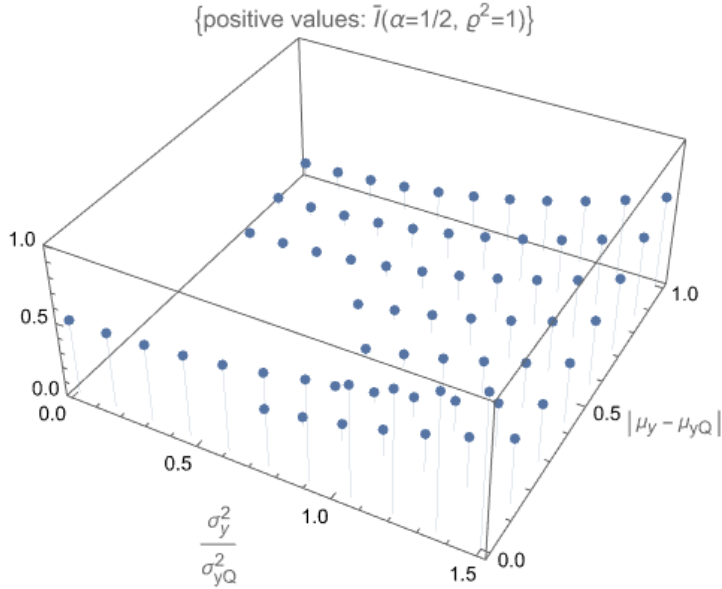

Figure 2. $\overline{I}_\alpha^{GM}(P_Y, Q_Y | \rho \to 1)$. It is $\alpha$ independent. The regions with absent points are those where $\min_{Q_Y} \neq P_Y$. See Eq. (D.20) that shows that as $|\mu_y - \mu_{y,Q}|$ increases the region of this behavior shrinks.

Figures 3.A-C display plots for the general case in Eq. (D.18) for a set of increasing $0 \leq \alpha \leq 1$ values. There are again some regions where $\min Q_Y \neq P_Y$, concentrated in regions where $\rho$ is relatively large and $(\mu_y - \mu_{y,Q})^2$ is relatively small. Otherwise $\min Q_Y \to P_Y$ as for Shannon VAEs.

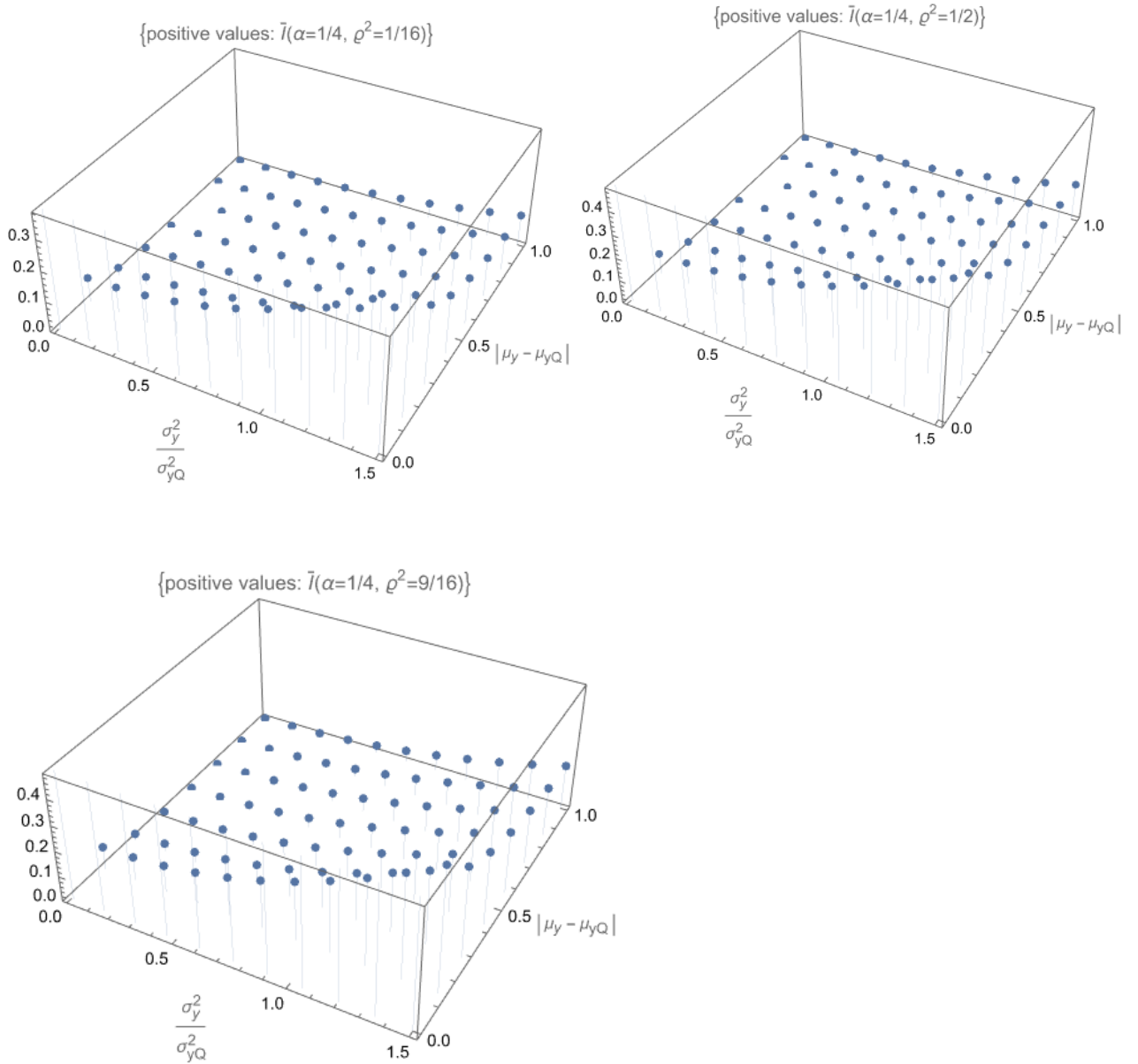

Figure 3.A. $\overline{I}_\alpha^{GM}(P_Y, Q_Y)$ for $\alpha = 1/4$ and various $\rho^2$ values where $\min_{Q_Y} \to P_Y$ is favored.

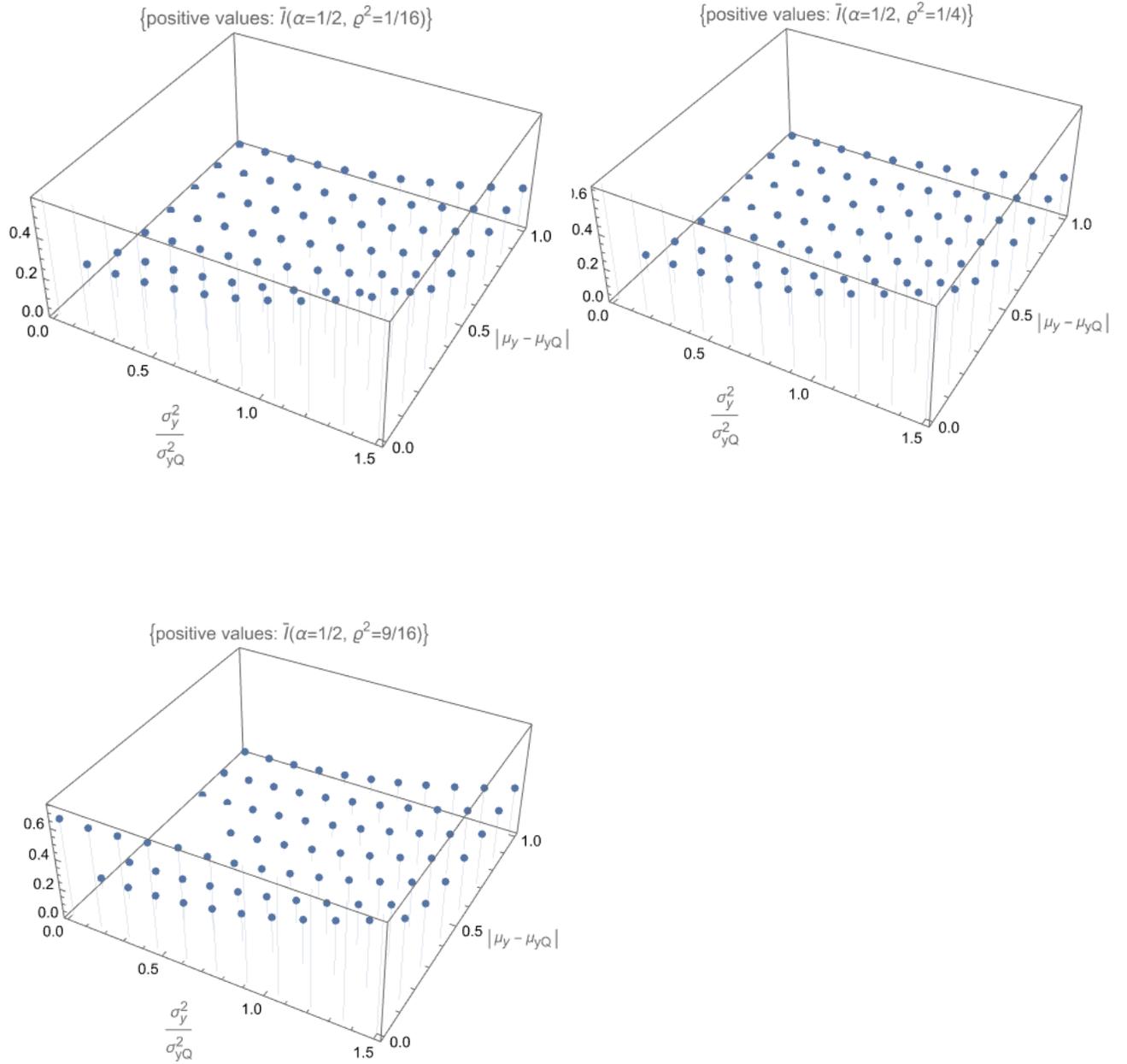

Figure 3.B. $\bar{I}_\alpha^{GM}(P_Y, Q_Y)$ for $\alpha = 1/2$ and various $\rho^2$ values where $\min_{Q_Y} \to P_Y$ is favored. There are some missing values indicating $\min_{Q_Y} \neq P_Y$. They occur more as $\rho^2 \to 1$ for $\sigma_y^2 / \sigma_{y,Q}^2 \leq 1$.

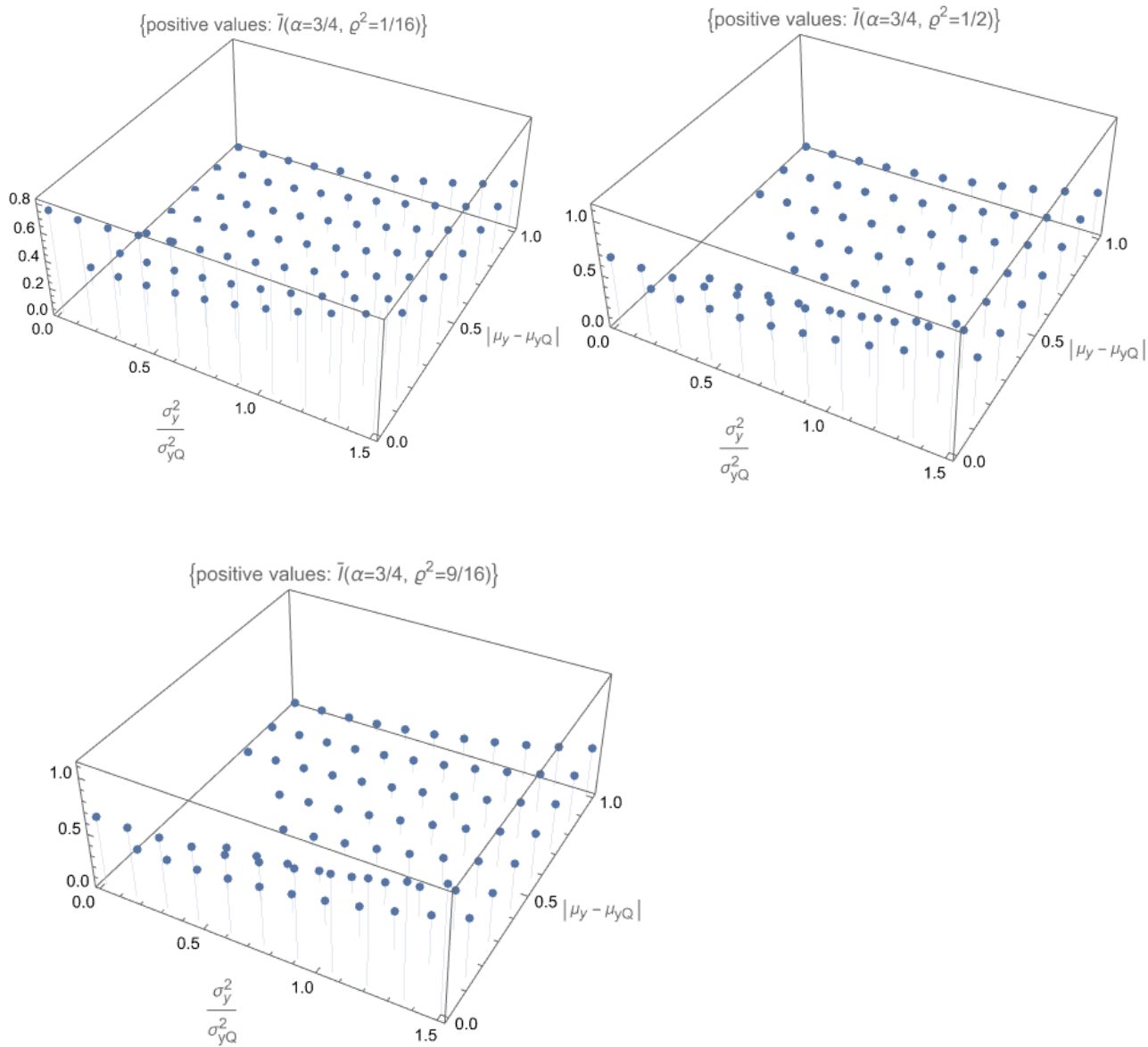

Figure 3.C. $\bar{I}_\alpha^{GM}(P_Y, Q_Y)$ for $\alpha = 3/4$ and various $\rho^2$ values where $\min_{Q_Y} \to P_Y$ is favored. There are some missing values indicating $\min_{Q_Y} \neq P_Y$. These occur more as $\rho^2 \to 1$ for $\sigma_y^2/\sigma_{y,Q}^2 \leq 1$.

For $\alpha > 1$ and $\sigma_{y,Q}^2/\sigma_y^2 \leq 1$ there is the possibility of failure of the second condition in Eq. (D.9) $\sigma_{\alpha,y,Q}^2/\sigma_y^2 = (1-\alpha)(1-\rho^2) + \alpha\, \sigma_{y,Q}^2/\sigma_y^2 \geq 0$. Figure 4 displays, for values that do not have this failure, results similar to the $\alpha \leq 1$ situation.

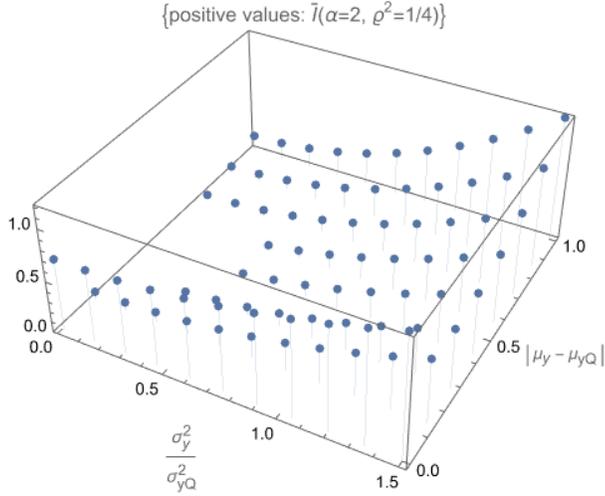

Figure 4. $\overline{I}_\alpha^{GM}(P_Y, Q_Y)$ for $\alpha > 1$, the missing values indicating $\min_{Q_Y} \neq P_Y$ are similar to those for $\alpha < 1$.

When increasing $\alpha$, avoiding the failure condition in Eq. (D.9), the minimization does indicate a broader range of $\min_{Q_Y} \neq P_Y$ as shown in Figure 5.

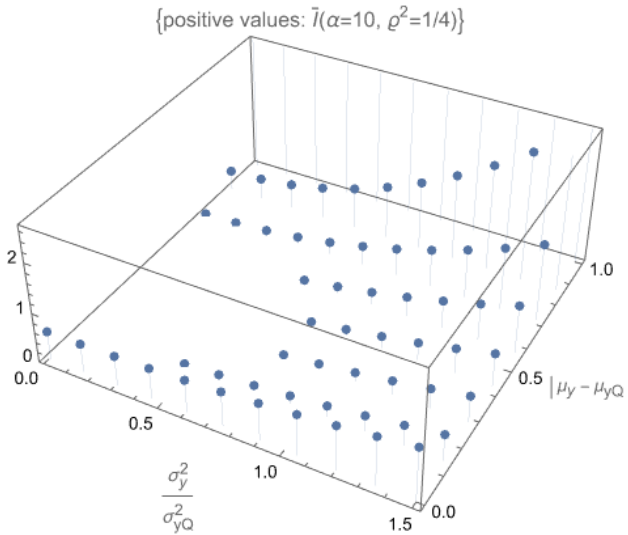

Figure 5. $\overline{I}_\alpha^{GM}(P_Y, Q_Y)$ for $\alpha = 10$ showing greater range of $\min_{Q_Y} \neq P_Y$.